\documentclass{article} 
\usepackage{iclr2023_conference,times}

\usepackage{amsmath,amsfonts,bm}

\def\eqref#1{equation~\ref{#1}}

\def\1{\bm{1}}

\DeclareMathAlphabet{\mathsfit}{\encodingdefault}{\sfdefault}{m}{sl}
\SetMathAlphabet{\mathsfit}{bold}{\encodingdefault}{\sfdefault}{bx}{n}

\DeclareMathOperator*{\argmax}{arg\,max}

\usepackage{hyperref}
\usepackage{graphicx}
\usepackage{url}
\usepackage{wrapfig}
\usepackage{graphics}
\usepackage{booktabs}
\usepackage{subcaption}
\usepackage{multirow}
\usepackage{bbm}
\usepackage{amsmath,amsfonts,bm}
\usepackage{makecell}
\usepackage{array}
\usepackage[para]{threeparttable}
\usepackage{booktabs}
\usepackage{tabularx}
\usepackage{xspace}

\usepackage[11pt]{moresize}

\newcolumntype{?}{!{\vrule width 1pt}}
\title{Guess the Instruction! \newline Flipped Learning Makes Language Models Stronger Zero-Shot Learners}

\author{Seonghyeon Ye{\textsuperscript{1}\thanks{~~Work done while interning at LG AI Research.}} \quad Doyoung Kim{\textsuperscript{1}}\quad Joel Jang{\textsuperscript{1}\footnotemark[1]}\quad Joongbo Shin{\textsuperscript{2}} \quad Minjoon Seo{\textsuperscript{1}} \\  \\
{\textsuperscript{1}}KAIST\quad{\textsuperscript{2}}LG AI Research\\ 
\texttt{\{seonghyeon.ye,ikevin98,joeljang,minjoon\}@kaist.ac.kr} \\
\texttt{jb.shin@lgresearch.ai} \\
}

\newcommand{\ours}{\textsc{Flipped}\xspace}
\newcommand{\training}{\textsc{Flipped Learning}\xspace}

\iclrfinalcopy 
\begin{document}

\maketitle
\begin{abstract}
Meta-training, which fine-tunes the language model (LM) on various downstream tasks by maximizing the likelihood of the target label given the \textit{task instruction} and input instance, has improved the zero-shot task generalization performance. However, meta-trained LMs still struggle to generalize to challenging tasks containing novel labels unseen during meta-training. In this paper, we propose \training, an alternative method of meta-training which trains the LM to generate the task instruction given the input instance and label. During inference, the LM trained with \training, referred to as \ours, selects the label option that is most likely to generate the task instruction. On 14 tasks of the BIG-bench benchmark, the 11B-sized \ours outperforms zero-shot T0-11B~\citep{sanh2021multitask} and even a 16 times larger 3-shot GPT-3 (175B)~\citep{brown2020language} on average by 8.4\% and 9.7\% points, respectively. \ours gives particularly large improvements on tasks with unseen labels, outperforming T0-11B by up to +20\% average F1 score. This indicates that the strong task generalization of \ours comes from improved generalization to novel labels. We release our code at \href{https://github.com/seonghyeonye/Flipped-Learning}{github.com/seonghyeonye/Flipped-Learning}.
\end{abstract}

\section{Introduction}
Large Language Models (LMs) pretrained on a vast amount of corpora are capable of solving various downstream tasks through instructions (task prompts) concatenated with the input instances without any task-specific fine-tuning \citep{brown2020language, rae2021scaling, chowdhery2022palm, zhang2022opt}. Previous work has shown that fine-tuning the LM on various downstream tasks by generating the correct answer given a prompted input (instruction and input), also referred to as \textit{meta-training}, leads to significant improvement in zero-shot task generalization \citep{sanh2021multitask, wei2021finetuned, wang2022benchmarking}.
However, \citet{webson2021prompt, min2022rethinking} show that LMs meta-trained through this standard approach are sensitive to different label words, implying that standard meta-trained LMs often fail to generalize to tasks that contain novel labels.

\begin{figure}[t]
\vspace{-2mm}
    \centering
    \includegraphics[width=\linewidth]{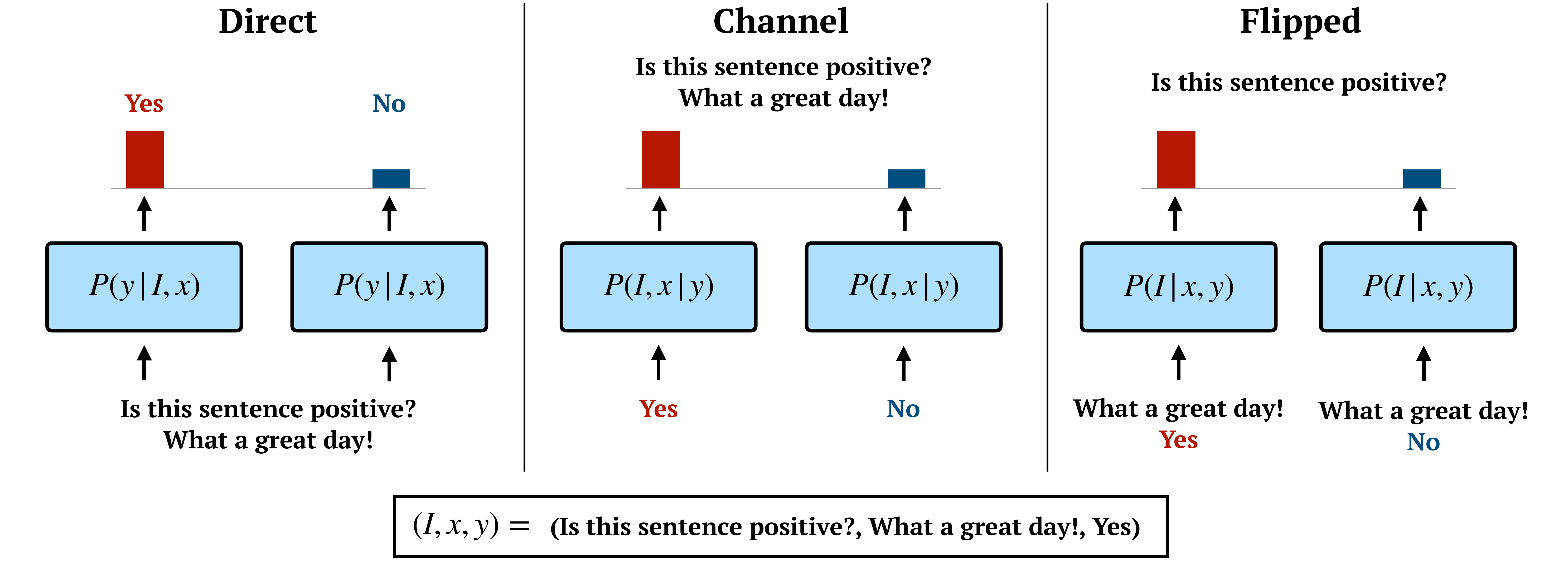}
    \caption{{Inference of \textsc{Direct}, \textsc{Channel} and \ours to select an appropriate label (Yes) from label options (Yes/No). \textsc{Direct}, which is the standard LM inference, computes the conditional probability of label given instruction+input. \textsc{Channel}, which is noisy channel inference, computes the conditional probability of instruction+input given label. Our \ours computes the conditional probability of instruction given input+label.}}
    \label{fig:fig_teaser} 
\vspace{-3mm}
\end{figure}

In this paper, we introduce an alternative meta-training method called \training that flips the task instruction and label space, training the underlying LM to generate the \textit{instruction} when given the input instance and label. This differs from the standard meta-training methods which train the LM to generate the label given instruction and input instance (\textsc{Direct}) or generate instruction and input instance given the label (\textsc{Channel}). Also, we add an unlikelihood loss for \training, making the LM not generate the task instruction for an incorrect label option.
During inference, the LM trained via \training, referred to as \ours, selects the label option that is most likely to generate the task instruction, as shown in Figure \ref{fig:fig_teaser}.



\begin{figure}[t]
    \centering
    \includegraphics[width=0.7\linewidth]{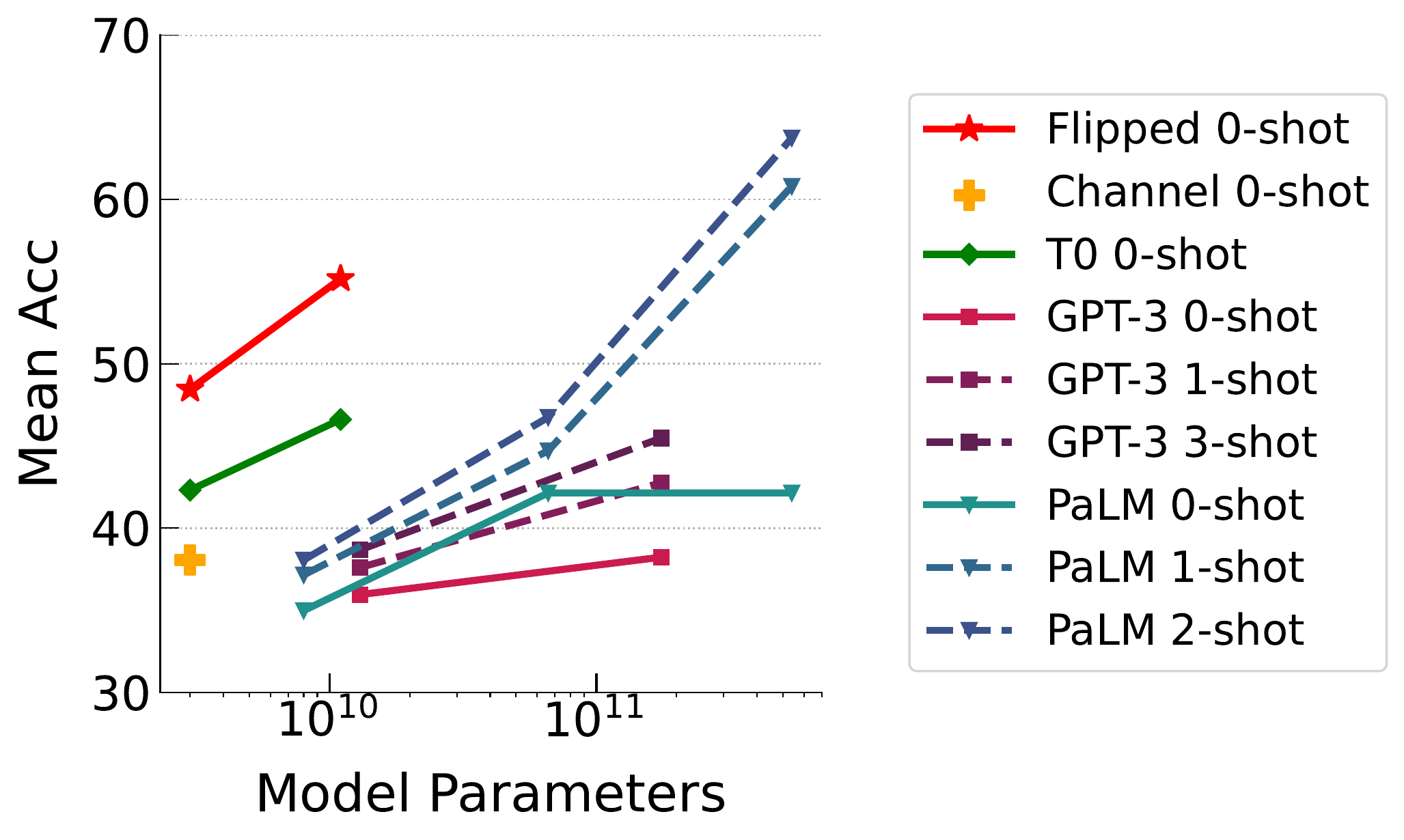}
    \caption{{Mean Accuracy on 14 datasets from the BIG-Bench benchmark. \ours shows the best performance among zero-shot LMs and even better performance than GPT-3 175B 3-shot.}}
    \label{fig:fig_big} 
\vspace{-5mm}
\end{figure}

To compare with an existing meta-trained LM T0~\citep{sanh2021multitask} trained by the \textsc{Direct} approach, we implement \ours by meta-training the T5~\citep{raffel2019exploring} model on 20 different datasets (around half of the datasets used to train T0) with only $\sim$5\% of training compute compared to T0.
Evaluation on 14 datasets from BIG-Bench~\citep{srivastava2022beyond} demonstrate that \ours is effective (Figure \ref{fig:fig_big}), not only showing state-of-the-art performance compared to all LMs regardless of size in the zero-shot setting, but also outperforming much larger GPT-3 175B (3-shot) by a significant margin, even without any demonstrations of the task (zero-shot). We also compare \ours with baseline models on 14 additional common English NLP tasks, further amplifying its effectiveness compared to previous methods and models.


We hypothesize that \ours shows strong zero-shot generalization ability on unseen tasks because of the improved generalization capability to unseen \textit{labels}. 
To test this hypothesis, we evaluate on various label pairs with different surface forms but with the same meaning (e.g. yes/no vs agree/disagree).
Results show \ours has up to +20\% average F1 score performance gap with T0-11B, indicating that \training indeed significantly improves label generalization capability. This hypothesis is further bolstered by the fact that the tasks that show significant performance improvement from the baselines among the 28 evaluation datasets are datasets with unseen labels during meta-training. Because \training conditions on the label instead of generating it, \training is likely to avoid label overfitting, resulting in improved label generalization, which consequently leads to better task generalization.

In summary, our contributions are as follows:
\begin{itemize}
    \item We propose \training, a novel meta-training method that computes the likelihood of the task instruction given the concatenation of input instance and label. By adding an unlikelihood loss, we make LMs generate the task instruction depending on the input instance-label correspondence. 
    
    \item On 14 datasets from the BIG-Bench benchmark, we show that 11B-sized \ours (LM trained through \training) outperforms not only meta-trained T0-11B by 8.4\% points on average, but also 16x larger 3-shot GPT-3 by 9.7\% points. When evaluating on 14 additional English NLP tasks, \ours outperforms all baseline models on average, further demonstrating the effectiveness of our proposed method.  
    
    
    \item We show that \ours is particularly effective on generalization to labels that are unseen during meta-training, outperforming T0-11B by up to 20\% average F1 score for novel label pairs.
\end{itemize}


\section{Related Work}
\subsection{Meta-training}
Prior work has shown that \textit{meta-training}, multitask fine-tuning on various downstream tasks with task instructions included, enables zero-shot task generalization \citep{sanh2021multitask, wei2021finetuned, wang2022benchmarking, mishra2021cross}. 
Specifically, \citet{sanh2021multitask, wang2022benchmarking} have shown that moderate-sized LMs can also generalize to unseen tasks through meta-training and the generalization performance improves by scaling the number of training tasks, the number of prompts per task, and the size of the LM. Based on this method, \citet{ouyang2022training} apply reinforcement learning with human feedback after meta-training to make better instruction-following LMs. To improve the task generalization performance of meta-trained LMs, \citet{lin2022unsupervised, ye2022retrieval} suggest using a retrieval-based framework. \citet{min-etal-2022-metaicl, chan2022few, chen2021meta} apply meta-training by using input-label pairs instead of task instructions.


\subsection{Noisy channel prompting}
When performing classification tasks, zero-shot LMs \citep{brown2020language, chowdhery2022palm} compute the conditional probability of the labels given input instances concatenated with instructions or demonstrations, referred to as \textit{direct prompting}. On the other hand, \textit{noisy channel prompting} reverts the input and the output space, making LMs generate every word in the input instances when conditioned on the label \citep{min-etal-2022-noisy, lazaridou2022internet}. Specifically, \citet{min-etal-2022-metaicl} apply noisy channel prompting during meta-training, optimizing the model to generate the input instance given the concatenation of demonstrations and the label. 
Motivated from \citet{min-etal-2022-metaicl}, we optimize the model to generate \textit{only} the task instruction while conditioning on the input and label (example shown in Figure \ref{fig:fig_teaser}). While \citet{honovich2022instruction, gupta2022improving} have similar intuition of guessing the instruction given input and label, they only do \textit{flipping} on either training or inference, not both.

\subsection{Label Generalization}
Previous work has shown that LMs are very sensitive to different label surface forms, indicating poor robustness.  
\citet{zhao2021calibrate} show that even 175B-sized GPT-3 suffers from high sensitivity and propose contextual calibration to solve this issue.
\citet{holtzman2021surface, shi2022nearest} define this problem as surface form competition and propose Domain Conditional Pointwise Mutual Information scoring or fuzzy verbalizers to mitigate this problem. 
For meta-training, \citet{webson2021prompt} analyze the effect of various label surface forms for a meta-trained LM and find that meta-trained LMs are more sensitive to label surface forms than different wordings of the prompt, which suggests that the meta-trained LMs \textit{overfit} to the label space provided during meta-training. This shows that meta-trained LMs cannot generalize to unseen label space, indicating poor \textit{label generalization}.

\section{\training}
In this section, we introduce \training, which trains the LM to compute the conditional probability of the task instruction given input instance and label (Figure \ref{fig:fig_teaser}). We first introduce notations and compare the difference between previous approaches (\textsc{Direct} and \textsc{Channel}) and our proposed method during inference (Section \ref{sec:instruct_channel}).
Next, we provide a detailed explanation of the training objective of \training during meta-training and explain the intuition for including an unlikelihood training loss in addition to the original loss (Section \ref{sec:unlikelihood}).

\subsection{Inference of Probabilistic LMs}
\label{sec:instruct_channel}
In this work, we focus on tasks with label options such as classification and multi-choice tasks for both meta-training and evaluation. For a given task $T = \{x, Y\}$ where $x$ is the input instance and $Y=\{y_1,...y_k\}$ is label option set, we convert the data instance into a prompted version $\{[I,x],L\}$. From $\{[I,x],L\}$, $[I,x]$ denotes the prompted input instance including natural language instruction $I$ and $L= \{l_1,...,l_k\}$ denotes the natural language label option set where $l_i = v_I(y_i)$ and $v_I$ is the verbalizer corresponding to $I$. The goal during inference is to select the correct $l_i$ from $L = \{l_1...l_k\}$ given $I$ and $x$.  

\textbf{\textsc{Direct}} method computes the conditional probability of the label given task instruction and input instance. During inference, it selects the label that leads to the highest conditional probability:
\begin{equation}
\argmax_{l_i} P(l_i|I,x) 
\end{equation}
This is the most common approach used for zero-shot inference of LMs \citep{brown2020language, chowdhery2022palm, sanh2021multitask, wei2021finetuned}.

\textbf{\textsc{Channel}} method \citep{min-etal-2022-noisy} computes the conditional probability of instruction and input instance given a label. Using Bayes' rule, the probability can be reparameterized as follows:
\begin{equation}
\argmax_{l_i} P(l_i|I,x) = \argmax_{l_i} \frac{P(I,x|l_i)P(l_i)}{P(I,x)} = \argmax_{l_i} P(I,x|l_i) 
\end{equation}
since $P(I,x)$ is independent from $l_i$ and $P(l_i)= \frac{1}{|L|}$; we assume the prior to be an uniform distribution for tasks with label options. 

\textbf{\training}, our proposed method, computes the conditional probability of the task instruction given an input instance and a label. Different from previous approaches, we separate $[I,x]$ into $I$ and $x$ and use Bayes' rule to reparameterize the conditional probability as follows: 
\begin{equation}
\argmax_{l_i} P(l_i|I,x) = \argmax_{l_i} \frac{P(I|x,l_i)P(l_i,x)}{P(I,x)} = \argmax_{l_i} P(I|x,l_i)P(l_i|x) \approx \argmax_{l_i} P(I|x,l_i) 
\end{equation}
where we assume $P(l_i|x)\approx \frac{1}{|L|}$ for simplicity. 
By considering $P(I|x,l_i)$, we allow the LM to put more focus on the task instruction.
The intuition of \training can be considered to be similar to generative question answering \citep{lewis2018generative} which generates the question given context and answer, but \training generates the task instruction for task generalization.

To compute $P(I|x,l_i)$ for \training, the prompted input $[I,x]$ should be separated into task instruction $I$ and input instance $x$. However, the prompted input might be sometimes an intermix of $I$ and $x$ rather than their sequential concatenation. To handle this, we follow \citet{raffel2019exploring}'s denoising objective using sentinel tokens where the portions representing task instruction $I$ are replaced by sentinel tokens and the LM learns to denoise the sentinel tokens by generating $I$.\footnote{We illustrate the denoising objective example in Appendix \ref{appen:denoise_ill}.}

\subsection{Meta-training using \training}
\label{sec:unlikelihood}
Next, we explain how we optimize the LM to utilize $P(I|x,l_i)$ which requires adding in an \textit{unlikelihood} loss during meta-training.
Given the sequence of task instruction $I=(I_1,..,I_T)$, we denote the LM loss function as follows:
\begin{equation}
L_{LM} = - \sum_{t=1}^T \log P(I_t | x, l_c, I_{<t})
\end{equation}
where $l_c$ corresponds to the correct label option. By minimizing this loss function, the LM learns to generate $I$ when given the correct label option and the input instance. 

\paragraph{Unlikelihood Loss} However, from preliminary experiments, we observe that meta-training the LM only on $L_{LM}$ results in ignoring the correspondence between the \textit{input instance} and \textit{label}: meta-trained LM generates task instruction $I$ regardless of the correspondence of the label option. We conjecture that this is a result of shortcut learning of large LMs \citep{du2022shortcut, min2022rethinking}.
To amplify the correspondence signal between the input instance and the correct label, we add an unlikelihood loss \citep{tam2021improving, liu2022few, welleck2019neural} during meta-training which can be denoted as follows: 
\begin{equation}
L_{UL} = - \sum_{t=1}^T \log (1- P(I_t | x, l_{c^{\prime}}, I_{<t}))
\end{equation}
where $l_{c^{\prime}}$ corresponds to an incorrect label option randomly sampled from the incorrect label option set $L_{C^{\prime}} = \{l|l \in L, l \ne l_c\}$. This unlikelihood loss term allows the LM to \textit{not} generate the task instruction if the label option does not correspond to the input instances. The final training objective of \training is the weighted sum of $L_{LM}$ and $L_{UL}$: 
\begin{equation}
L = L_{LM} + \lambda L_{UL}
\end{equation}
where $\lambda$ is a hyperparameter. By optimizing both likelihood and unlikelihood objectives, the LM is optimized to generate the instruction when given the correct label and not generate the instruction when given the incorrect label, strengthening the correspondence between the input instance and the correct label.



\section{Experimental Setup}
\paragraph{Training} For meta-training, we utilize the subset of T0 \citep{sanh2021multitask} meta-training datasets: 4 task clusters (sentiment classification, paraphrase detection, topic classification, multi-choice QA), which are 20 datasets in total. We only train on tasks with label options and exclude tasks such as free-form generation because \training requires label options for unlikelihood training on incorrect label options. We provide detailed training configurations in Appendix \ref{appen:config} and the full list of training datasets in Appendix \ref{appen:train_datasets}.

\paragraph{Evaluation} Following the evaluation setting of \citet{sanh2021multitask}, we first measure unseen task generalization performance on 14 tasks of BIG-bench which contain challenging and various tasks that are unseen during meta-training. For each of the BIG-bench tasks, we report the accuracy of a single instruction for each task following the convention of past work~\citep{sanh2021multitask, lin2022unsupervised}. Furthermore, we additionally evaluate on 14 English NLP unseen tasks, consisting of 7 classification and 7 multi-choice datasets, also following the setting of \citet{sanh2021multitask, lin2022unsupervised}. For evaluation metric, we use Macro-F1\footnote{Macro-F1 is more appropriate for imbalanced classification than accuracy.} for classification and accuracy for multi-choice tasks, following \citet{min-etal-2022-metaicl, min2022rethinking}. We also report the average standard deviation among different evaluation instructions, indicating the robustness of different wordings of the evaluation instruction (the lower, the better). For analysis of label generalization of classification tasks, we evaluate on 5 datasets: 2 seen datasets during meta-training (IMDB, PAWS) and 3 unseen datasets (RTE, CB, WiC). We provide the full list of evaluation datasets in Appendix \ref{append:eval_datasets} and more details on the evaluation setting are specified in Appendix \ref{appen:evaluation}.

\paragraph{Baselines}
We evaluate several baselines to observe the effectiveness of \training: (1) \textsc{T0-3B}, a 3B-sized meta-trained LM by \citet{sanh2021multitask}, (2) \textsc{Direct}, a 3B-sized meta-trained LM using the same language modeling objective (standard method) of \textsc{T0-3B}, but with our training configurations, 
(3) \textsc{Channel}, a 3B-sized meta-trained LM using noisy channel language modeling objective, 
(4) \ours-3B, a 3B-sized LM meta-trained through our proposed \training, (5) \textsc{T0-11B}, a larger meta-trained LM of \citet{sanh2021multitask}, (6) \ours-11B, a larger meta-trained LM trained through \training, (7) GPT-3 \citep{brown2020language}, 175B sized pretrained LM, (8) PaLM \citep{chowdhery2022palm}, 540B sized pretrained LM. Note that \textsc{Direct}, \textsc{Channel}, and \ours is meta-trained on the same number of datasets and training steps.

\begin{table}[]
\small\addtolength{\tabcolsep}{-1pt}
\centering
\fontsize{8}{11}\selectfont
\begin{tabular*}{1\columnwidth}{l|cccc|cc|cc||cc}
      & \multicolumn{8}{c||}{\textit{Zero-shot}}                                                                                             & \multicolumn{2}{c}{\textit{Few-shot}} \\ \midrule
\multirowcell{2}{Dataset (metric)}         & \textsc{T0}          & \textsc{Dir.}         & \textsc{Chan.}        & \textsc{Flip.}            & \textsc{T0}  & \textsc{Flip.} & \textsc{GPT-3}  & \textsc{PaLM}           & \textsc{GPT-3} (3)   & \textsc{PaLM} (1)  \\
      & 3B             & 3B             & 3B   & 3B    & 11B   & 11B               & 175B           & 540B           & 175B          & 540B         \\ \midrule
Known Un.     & 47.83 & 63.04 & 52.17 & 71.74 & 58.70    & \textbf{86.96}  & 60.87          & 56.52          & 50.00         & 67.39        \\
Logic Grid    & 41.10 & 35.90 & 30.90 & 41.70 & 38.30  & \textbf{42.50} & 31.20 & 32.10          & 31.10         & 42.20        \\
Strategy.     & 52.79          & 53.28          & 53.01          & 53.19          & 52.75     & 53.23           & 52.30          & \textbf{64.00} & 57.10         & 69.00        \\
Hindu Kn.     & 25.71          & 50.29         & 16.57          & 47.43          & 29.71 & 52.57               & 32.57          & \textbf{56.00} & 58.29         & 94.86        \\
Movie D.      & 52.85          & 47.15          & 51.06          & 47.93          & \textbf{53.69}  & 48.49     & 51.40          & 49.10          & 49.40         & 57.20        \\
Code D.       & 46.67& 33.33          & \textbf{71.67}          & 45.00          & 43.33       & 60.00         & 31.67          & 25.00          & 31.67         & 61.67        \\
Concept       & 45.52          & 58.14          & 35.67         & 61.64          & \textbf{69.29}  & 64.93     & 26.78          & 59.26          & 35.75         & 80.02        \\
Language      & 14.84          & 22.01 & 11.55      & 19.01          & 20.20    & \textbf{26.87}          & 15.90          & 20.10          & 10.90         & 37.30        \\
Vitamin       & 58.89          & 63.83          &    15.73           & 57.07          & 64.73 & \textbf{65.57}   & 12.30          & 14.10          & 52.70         & 70.40        \\
Syllogism     & \textbf{52.94} & 49.85          & 50.43          & 50.56          & 51.81     & 50.39           & 50.50          & 49.90          & 52.80         & 52.20        \\
Misconcept.   & 50.23         & 50.23          & 47.79         & 46.58          & 50.00    & \textbf{54.34}   & 47.95          & 47.49          & 60.27         & 77.63        \\
Logical       & 46.64          & 38.06          & 25.73         & 59.82 & 54.86    & \textbf{64.56}            & 23.42          & 24.22          & 33.93         & 34.42        \\
Winowhy       & 44.29          & 44.33          & \textbf{55.36}          & 53.33 & 52.11    & 55.08            & 51.50          & 45.30          & 56.50       & 47.50        \\
Novel Con.    & 15.63          & 3.13           &  15.63              & 25.00          & 15.63     &\textbf{46.88}           & \textbf{46.88} & \textbf{46.88} & 56.25         & 59.38        \\ \midrule
BIG-bench AVG & 42.56          & 43.75          & 38.07         & 48.57 & 46.79            & \textbf{55.17}    & 38.23          & 42.14          & 45.48         & 60.80        \\ \midrule
\end{tabular*}
\caption{{Task generalization performance on 14 BIG-bench tasks. \textsc{Dir.} denotes \textsc{Direct}, \textsc{Chan.} denotes \textsc{Channel}, and \textsc{Flip.} denotes \ours. Parentheses in the \textit{Few-shot} column denote the number of shots. \ours performs the best on average for zero-shot setting.}}
\vspace{-5mm}
\label{table:main}
\end{table}

\section{Experimental Results}
In this section, we evaluate the effectiveness of \ours compared to various baselines. For task generalization performance, we evaluate on 14 tasks of BIG-bench and 14 common English NLP tasks (Section \ref{sec:result_main}).
For analysis of \ours, we evaluate on multiple label options with the same meaning but with different surface forms (Section \ref{sec:result_label}). 

\subsection{Main Results}
\label{sec:result_main}

\paragraph{\textsc{Direct} outperforms T0-3B.}
Our implementation of \textsc{Direct} outperforms \textsc{T0-3B} significantly despite using only half of the datasets used to train T0 (20 out of 38), indicating competitive performance of our baselines.
Note that \textsc{T0-3B} and \textsc{Direct} are trained on the same training objective, with the only difference in training configurations, showing that our training setting is more optimal to evaluate the task generalization performance of meta-trained LMs.
We conjecture the significant performance improvement to come from 3 potential factors: (1) We train for about 5\% token updates compared to \citet{sanh2021multitask}, implying that \textsc{Direct} avoids overfitting to the training dataset (2) \textsc{Direct} includes the EOS (end-of-sequence) token during training and evaluation unlike \textsc{T0-3B}, (3) We do not use sequence-packing during meta-training. Our \textsc{Direct} baseline shows that the task generalization performance from \citet{sanh2021multitask} might have been underestimated
\footnote{Although one might think that the other 18 training datasets of T0 might have had a negative impact on unseen tasks, we observe that training on whole datasets also shows similar results to \textsc{Direct} from preliminary experiments, implying that the performance improvement comes from the 3 potential factors mentioned above.}.

\begin{table}[]
\centering
\small\addtolength{\tabcolsep}{-1pt}
\centering
\fontsize{8}{11}\selectfont
\begin{tabular}{l|cccc|cc|c}
\toprule
\multirowcell{2}{Dataset (metric)}         & \textsc{T0}          & \textsc{Dir.}         & \textsc{Chan.}       & \textsc{Flip.}            & \textsc{T0}  & \textsc{Flip.}             & \textsc{GPT-3}  \\
      & 3B             & 3B             & 3B   & 3B    & 11B    & 11B              & 175B      \\ \midrule
RTE (F1)           & 61.89          & 72.83          & 36.62          & 71.03 & \textbf{80.91} &  72.20     & 40.68 \\
CB (F1)           & 30.94          & 49.81          & 22.35          & 52.27          & 53.82   & \textbf{61.51}    & 29.72   \\
ANLI R1 (F1)       & 24.39          & 30.17          & 21.30         & 33.92          & 34.72 & \textbf{34.93}      & 20.90  \\
ANLI R2 (F1)       & 23.73          & 28.23          & 21.44          & \textbf{32.62} & 31.25   & 32.59             & 22.50  \\
ANLI R3 (F1)       & 23.45          & 30.41          & 22.50          & 34.65 & 33.84     & \textbf{34.77}           & 23.77  \\
WSC (F1)           & 54.64          & 50.35          & 46.38          & 52.82          & \textbf{58.36}    & 49.88   & 26.24  \\
WiC (F1)           & 38.53          & 36.42          & 38.69          & 37.36          & \textbf{51.64}& 39.26               & 45.36 \\
COPA          & 75.88          & 89.63          & 50.13          & 89.88          & \textbf{91.50}  & 90.75             & 91.00 \\
Hellaswag     & 27.43          & 31.61          & 20.82          & 41.64          & 33.05       & 41.97         & \textbf{78.90}  \\
StoryCloze    & 84.03          & 94.24          & 57.84          & 95.88 & 92.40      & \textbf{96.12}          & 83.20          \\
Winogrande    & 50.97          & 55.96          & 50.99          & 58.56          & 59.94         & 66.57       & \textbf{70.20}  \\
PIQA          & 56.63          & 62.60          & 47.08          & 67.32          & 67.67         & 71.65       & \textbf{81.00}  \\
ARC-Chall     & 51.10          & 49.30          & 29.23         & 49.63          & 56.99     & \textbf{64.62}           & 51.40  \\
OpenbookQA    & 42.66          & 54.00          & 38.57          & 62.11          & 59.11         & \textbf{72.54}       & 68.80 \\ \midrule
En NLP AVG    & 46.16          & 52.54          & 36.00        & 55.69          & 57.51 &  \textbf{59.24}      & 52.41  \\
En NLP STD ($\downarrow$)    & 4.74           & 4.36           & 4.58          & 3.29  & 5.24 & \textbf{3.11} & -   \\ \bottomrule
\end{tabular}
\caption{{Zero-shot task generalization performance on 14 English NLP tasks consisted of 7 classification and 7 multi-choice tasks. 11B-sized \ours (\textsc{Flip.}) shows the best performance on average and also shows the best robustness to different evaluation instructions (lower STD).}}
\vspace{-5mm}
\label{table:nlp}
\end{table}

\paragraph{\textsc{Channel} is not effective for task generalization.}
\textsc{Channel} method largely underperforms \textsc{Direct} and \textsc{T0}, showing close to random guessing performance for many unseen tasks. This result is consistent with that of \citet{min-etal-2022-metaicl} in that instructions on zero-shot (not few-shot) setting \textit{worsen} the task generalization performance for \textsc{Channel} method unlike the \textsc{Direct} method. 
We conjecture that because prompted inputs are mostly question-answer formats, it is unnatural for \textsc{Channel} to generate a question-like instruction given only an answer even after meta-training.
\paragraph{\ours outperforms baselines.}
For the 14 BIG-bench tasks of Table \ref{table:main}, \ours-3B significantly outperforms all meta-trained models with the same model size: +6.01\% mean accuracy compared to T0-3B and +4.82\% mean accuracy compared to \textsc{Direct}. \ours-3B also outperforms 4x times larger meta-trained T0-11B on average by +1.78\% points. This result is significant considering that the effect of scaling law is strong for zero-shot generalization of meta-trained models~\citep{wei2021finetuned, sanh2021multitask, wei2022emergent}. \ours-11B even shows better performance, outperforming T0-11B on average by +8.38\% points. Compared to even larger pretrained LMs evaluated in a few-shot setting, \ours-11B outperforms 3-shot GPT-3 which is 16x larger by 9.69\% points on average. When compared to 1-shot PaLM which is 50x larger, \ours outperforms on 4 tasks out of the 14 tasks. This shows that \ours is effective for generalizing to unseen tasks that are challenging, resulting in the best performance on the zero-shot setting even when compared to LMs with much larger sizes. 

For the 14 common English NLP tasks which are consisted of 7 classification and 7 multi-choice tasks shown in Table \ref{table:nlp}, \ours-3B outperforms baseline models with the same model size (\textsc{T0-3B}, \textsc{Direct}, \textsc{Channel}) on task generalization performance by a significant margin, largely reducing the gap between T0-11B. \ours-11B shows the best performance on average, outperforming T0-11B by 1.73\% points. Also, \ours shows the lowest standard deviation among multiple different evaluation instructions compared to other meta-trained baseline models, including T0-11B. This indicates that \ours is not only effective for zero-shot task generalization but also \textit{robust} to different surface forms of the instruction.

\subsection{Analysis of \ours}
\label{sec:result_label}
\paragraph{\ours significantly outperforms baselines for tasks with unseen labels.}

We first analyze the tasks that \ours outperforms baseline models and identify a clear correlation: \ours especially shows strong performance on tasks that contain many label options unseen during meta-training. For RTE, WSC, WiC datasets, which are consisted of \textit{seen} label options (yes/no), direct meta-trained LMs (\textsc{Direct}, \textsc{T0}) show strong performance as shown in Table \ref{table:nlp}. On the other hand, \ours shows strong performance on CB and ANLI datasets that contain an \textit{unseen} label (e.g. maybe). This can be seen as a result of effective label generalization of \ours; the prediction of \textsc{T0} and \textsc{Direct} is largely biased to label options that are seen during training (yes/no) while \ours makes balanced predictions (yes/no/maybe). Although the calibration method of \citet{zhao2021calibrate} is known to mitigate the prediction bias of LMs, we find that calibration worsens the performance of meta-trained LMs (Appendix \ref{appen:calibrate}). 

\begin{figure*}[ht!]
\centering
    \begin{subfigure}[b]{0.25\textwidth}
    \includegraphics[width=\textwidth]{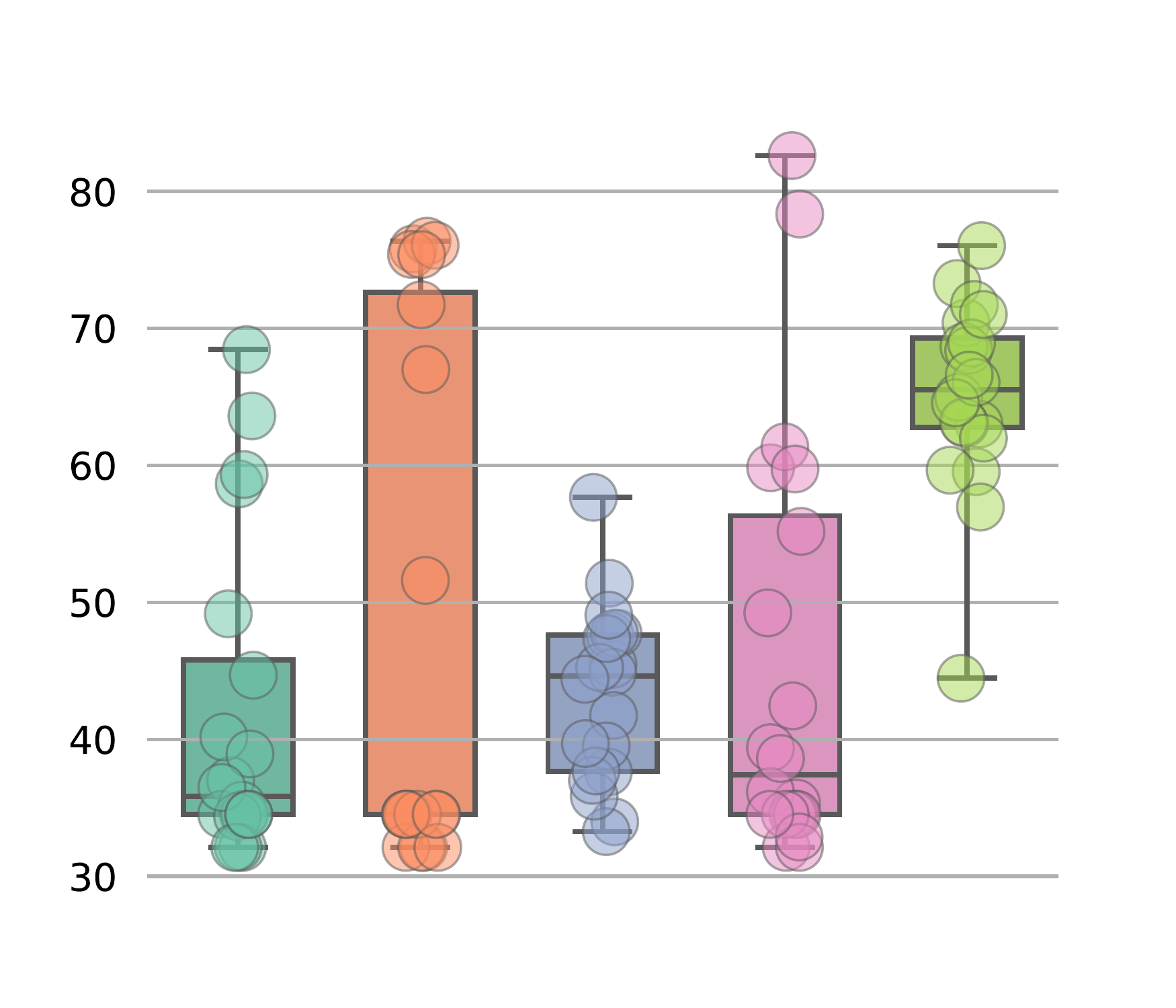}
    \caption{\small RTE}
    \label{fig:rte_label}
    \end{subfigure}
    \begin{subfigure}[b]{0.25\textwidth}
    \includegraphics[width=\textwidth]{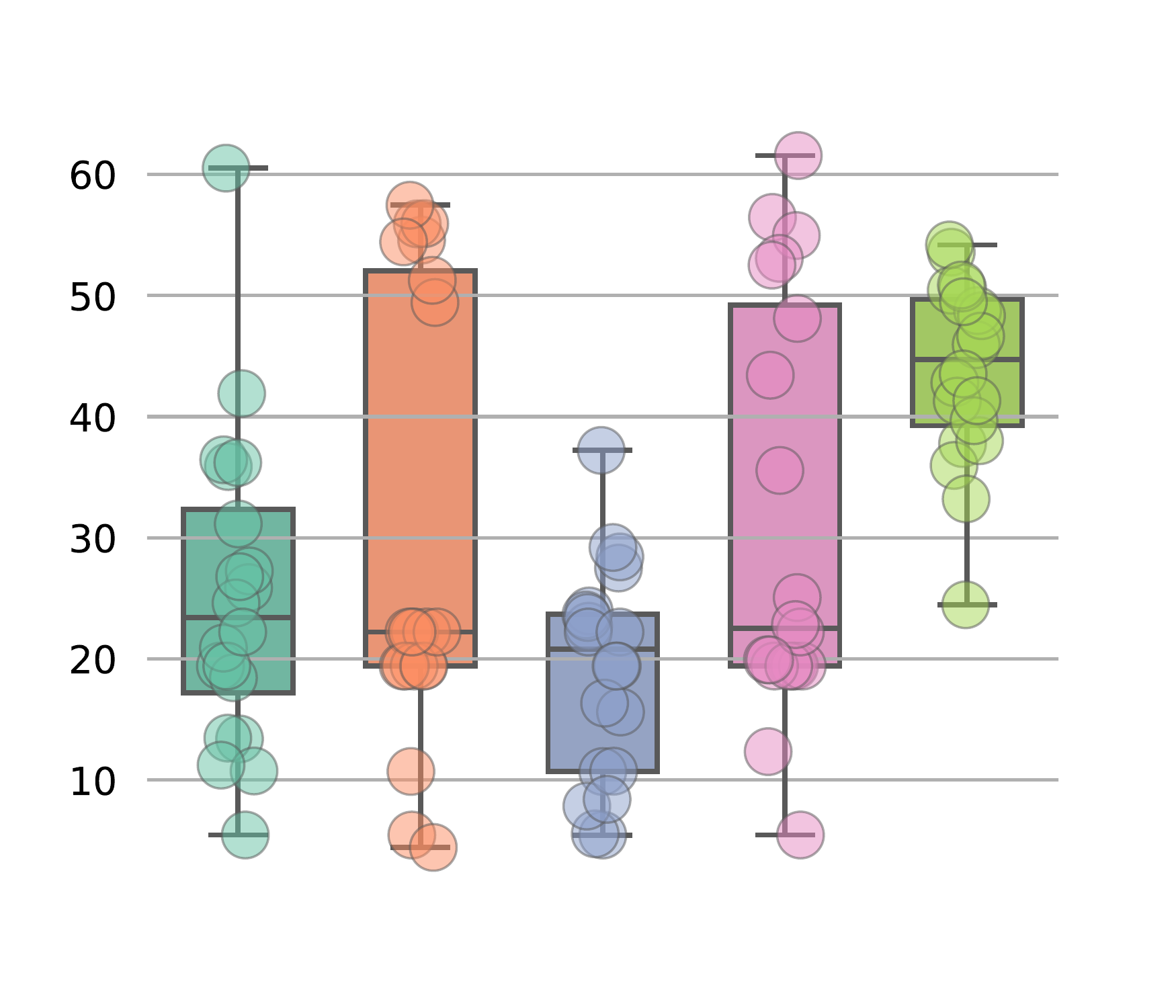}
    \caption{\small CB}
    \label{fig:cb_label}
    \end{subfigure}
    \begin{subfigure}[b]{0.25\textwidth}
    \includegraphics[width=\textwidth]{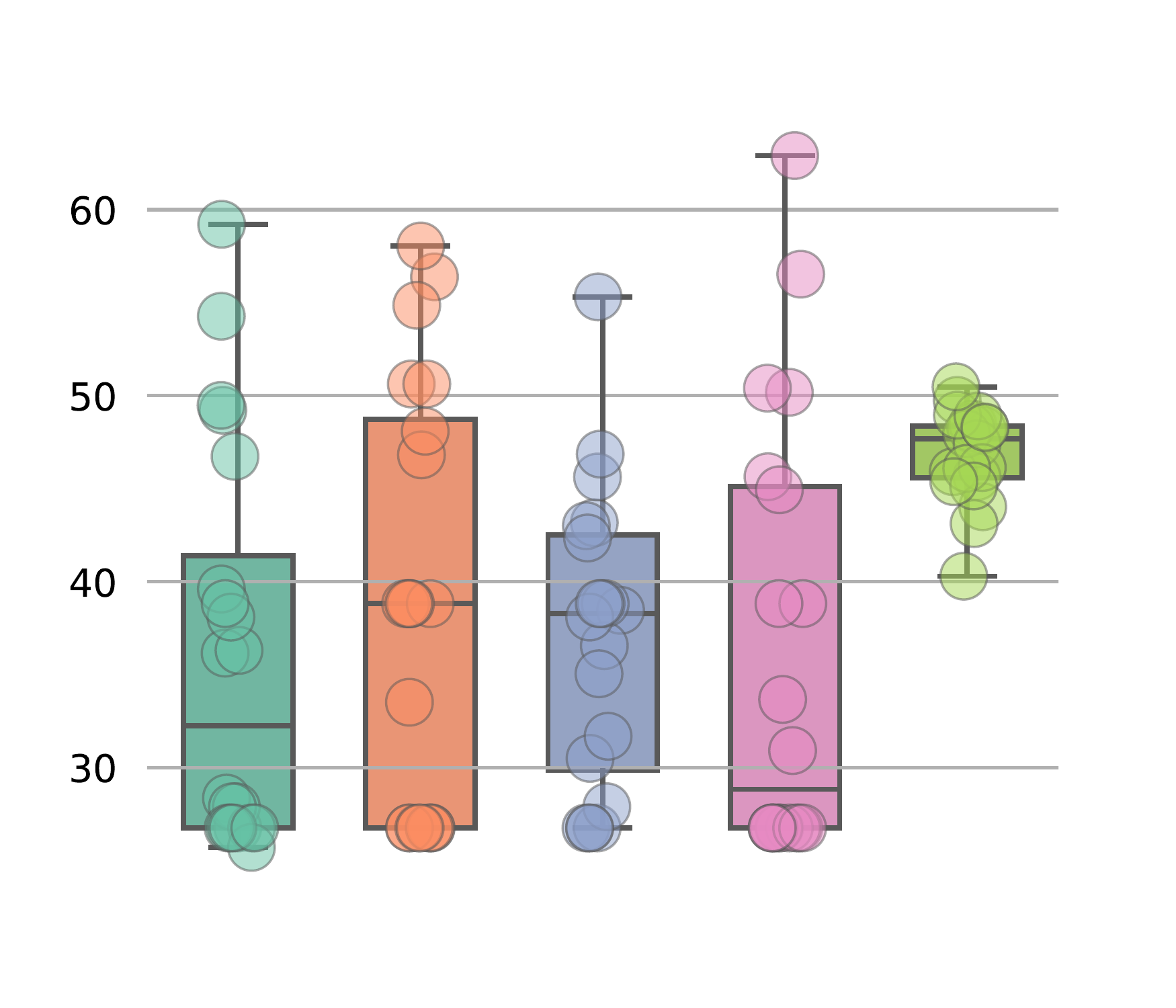}
    \caption{\small WSC}
    \label{fig:wsc_label}
    \end{subfigure}
    \begin{subfigure}[b]{0.25\textwidth}
    \includegraphics[width=\textwidth]{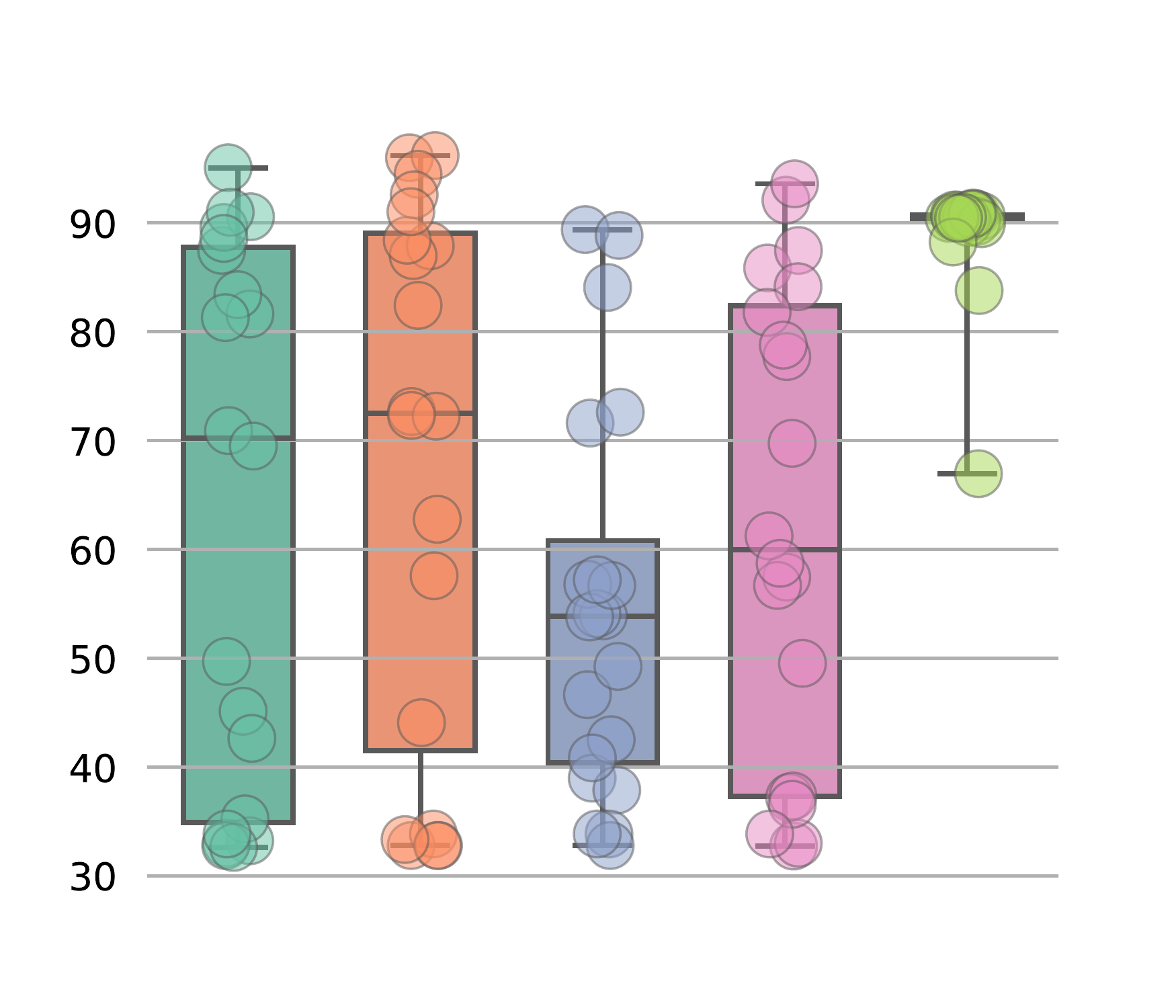}
    \caption{\small IMDB}
    \label{fig:wsc_label}
    \end{subfigure}
    \begin{subfigure}[b]{0.25\textwidth}
    \includegraphics[width=\textwidth]{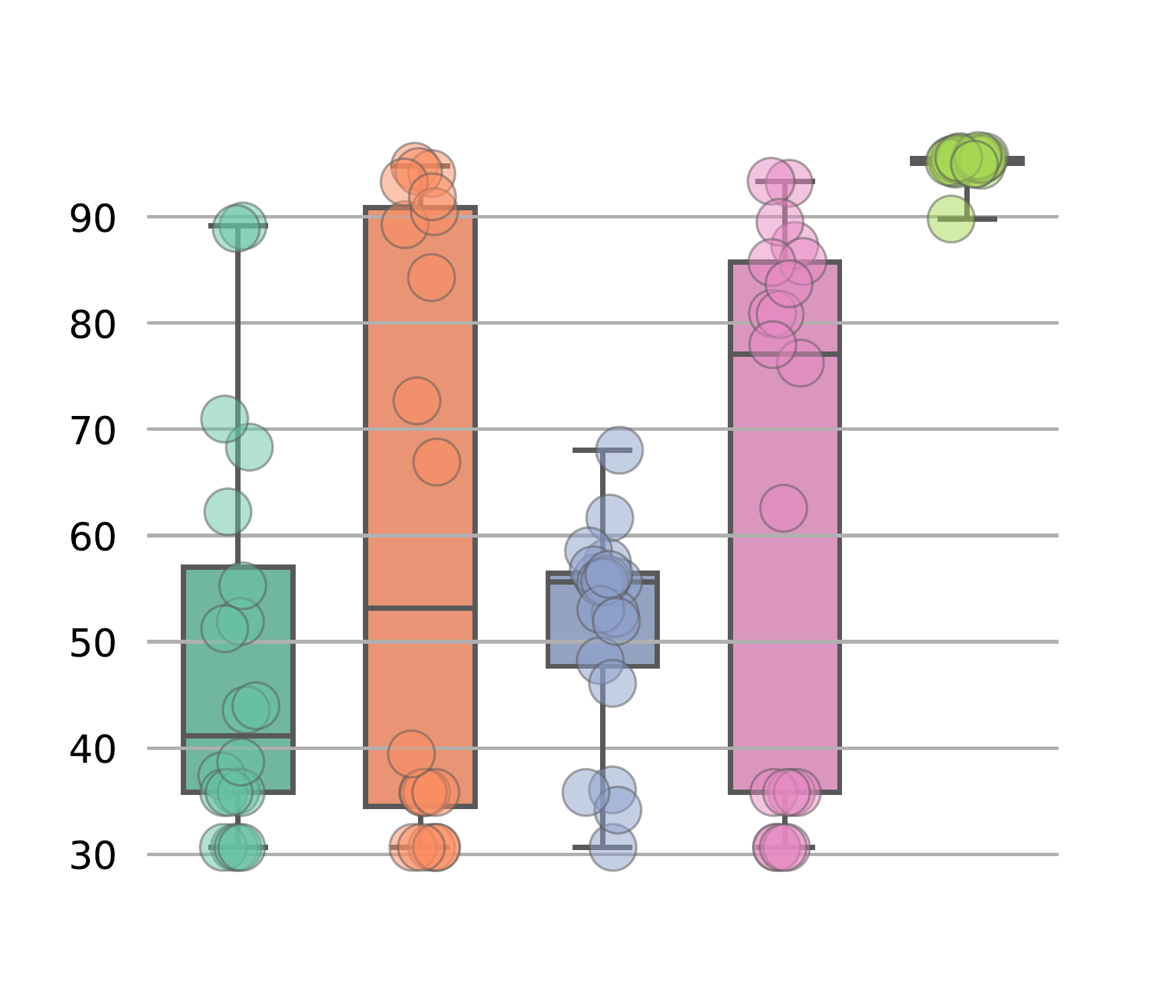}
    \caption{\small PAWS}
    \label{fig:wsc_label}
    \end{subfigure}
    \begin{subfigure}[b]{0.8\textwidth}
    \vspace{1mm}
    \includegraphics[width=\textwidth]{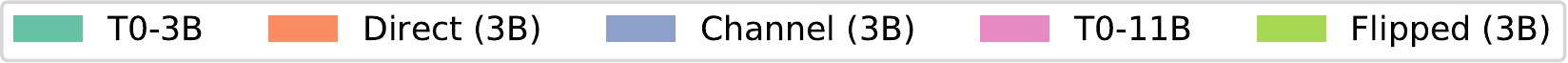}
    \end{subfigure}
\caption{{Label generalization performance on 3 unseen and 2 seen datasets during meta-training. We evaluate on 20 different label pairs including many unseen labels. Result shows that \ours significantly outperforms other baseline models.}}
\label{fig:label_generalize}
\vspace{-5mm}
\end{figure*}
For multi-choice tasks in Table \ref{table:nlp}, which are more likely to contain many unseen labels because they have different label options for every data instance, \ours outperforms meta-trained LMs with the same size for most tasks. Moreover, \ours outperforms T0-11B on BIG-bench (Table \ref{table:main}) which is mostly consisted of unseen labels.
From these findings, we can hypothesize that the strong zero-shot task generalization of \ours is likely to come from its strong label generalization capability. 

\paragraph{\ours generalizes to unseen labels that are semantically the same.}

To further test the above hypothesis, we analyze the label generalization performance of \ours compared to other baseline models by varying the surface form the label options (e.g. yes/no vs agree/disagree) for 5 classification datasets: 3 datasets (RTE, CB, WSC) for unseen tasks, and 2 datasets (IMDB, PAWS) for seen tasks during meta-training. We vary the label options to 20 different pairs that have the same meaning but different surface forms including the original labels.\footnote{We collected synonyms from \url{https://www.thesaurus.com}. We provide the full list of 20 label options in Appendix \ref{appen:label_20}.}

Figure \ref{fig:label_generalize} shows the label generalization performance of \textsc{T0-3B}, \textsc{Direct}, \textsc{T0-11B} and \ours-3B. For unseen tasks, \ours outperforms \textsc{T0-3B} by (+23.37\%, +18.78\%, +10.92\%), outperforms \textsc{Direct} by (+16.42\%, +13.46\%, +7.82\%), and outperforms \textsc{Channel} by (+21.88\%, 24.84\%, 9.93\%) average F1 score on (RTE, CB, WSC) respectively. Even when compared with a 4x times larger meta-trained LM (\textsc{T0-11B}), \ours outperforms by (+19.72\%, +12.32\%, +10.81\%) average F1 score for (RTE, CB, WSC) respectively. This shows that \ours can generalize to various novel labels, which is what even larger meta-trained LMs trained through direct prompting cannot do. Although baseline models outperform \ours for best accuracy among different label pairs, this is mostly when the label is seen during meta-training (e.g. yes/no). The result of Figure \ref{fig:label_generalize} also indicates that the classification tasks evaluation setting of \citet{sanh2021multitask} overestimates the true generalization ability of LMs because \citet{sanh2021multitask} mostly evaluate unseen target tasks on labels that are \textit{seen} during meta-training (yes/no), which is not guaranteed for a \textit{true} zero-shot generalization scenario. 

Aligned with the experiments on the 3 unseen tasks, \ours further outperforms baselines on the 2 \textit{seen} tasks during meta-training by a significant margin: (+25.55\%, +46.68\%) for \textsc{T0-3B}, (+20.74\%, +34.66\%) for \textsc{Direct}, (+34.12\%, +43.74\%) for \textsc{Channel}, and (+26.63\%, +31.91\%) for \textsc{T0-11B} on (IMDB, PAWS). 
This further bolsters the hypothesis that standard meta-training leads to label overfitting, especially for seen tasks and \training avoids this by conditioning on the label option instead of generating it. 


\section{Ablation Studies}
\label{sec:ablation}
In this section, we analyze the effect of unlikelihood training. Also, we vary the number of meta-training datasets of \ours to analyze the effect of the number of datasets on task generalization. We evaluate on 14 English NLP tasks and report average F1 score on 7 classification tasks and mean accuracy on 7 multi-choices tasks respectively. 

\begin{figure}[t]
    \centering
    \begin{subfigure}[b]{0.4\textwidth}
    \includegraphics[width=\textwidth]{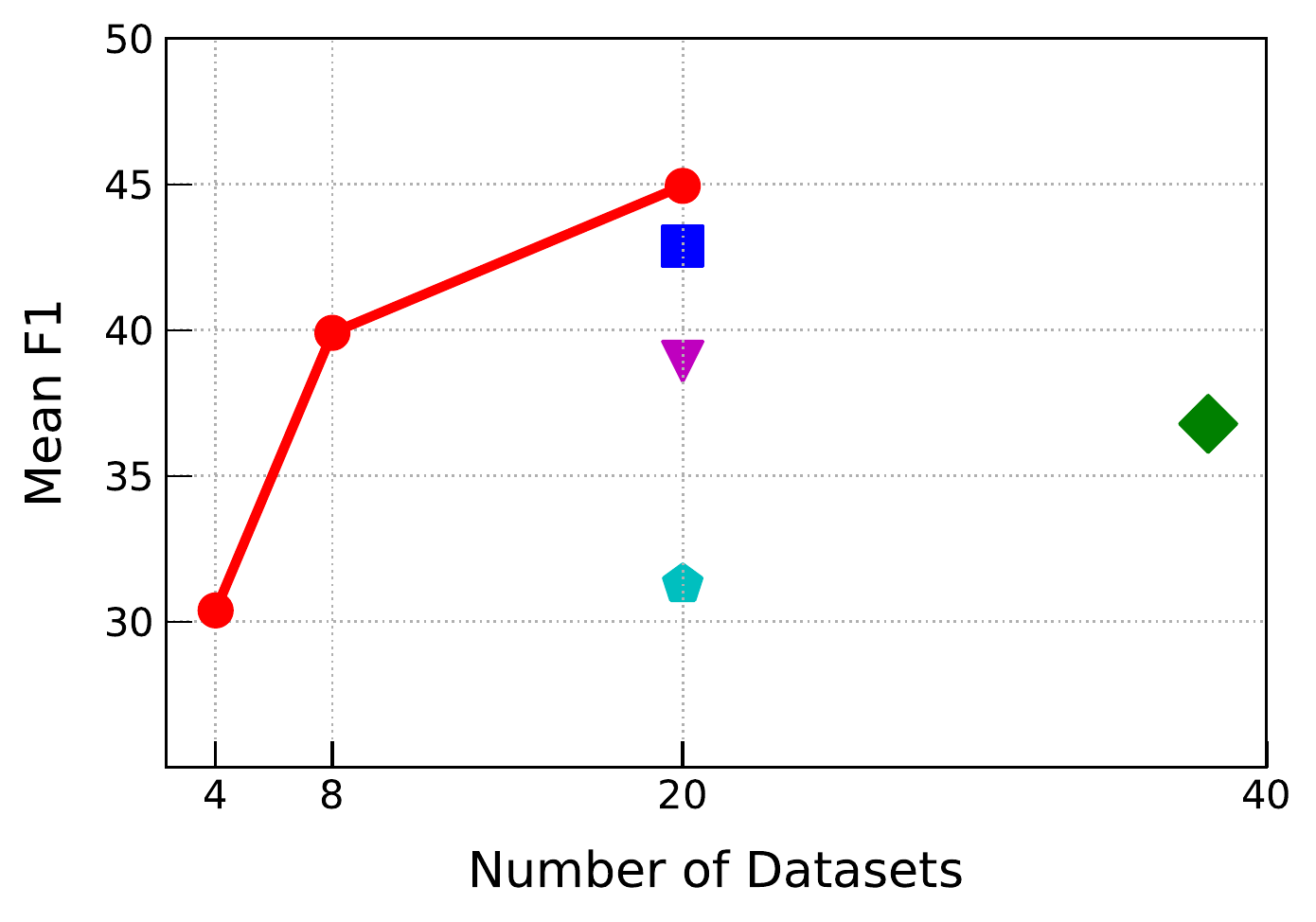}
    \label{fig:data_inst}
    \end{subfigure}
    \hspace{5mm}
    \begin{subfigure}[b]{0.4\textwidth}
    \includegraphics[width=\textwidth]{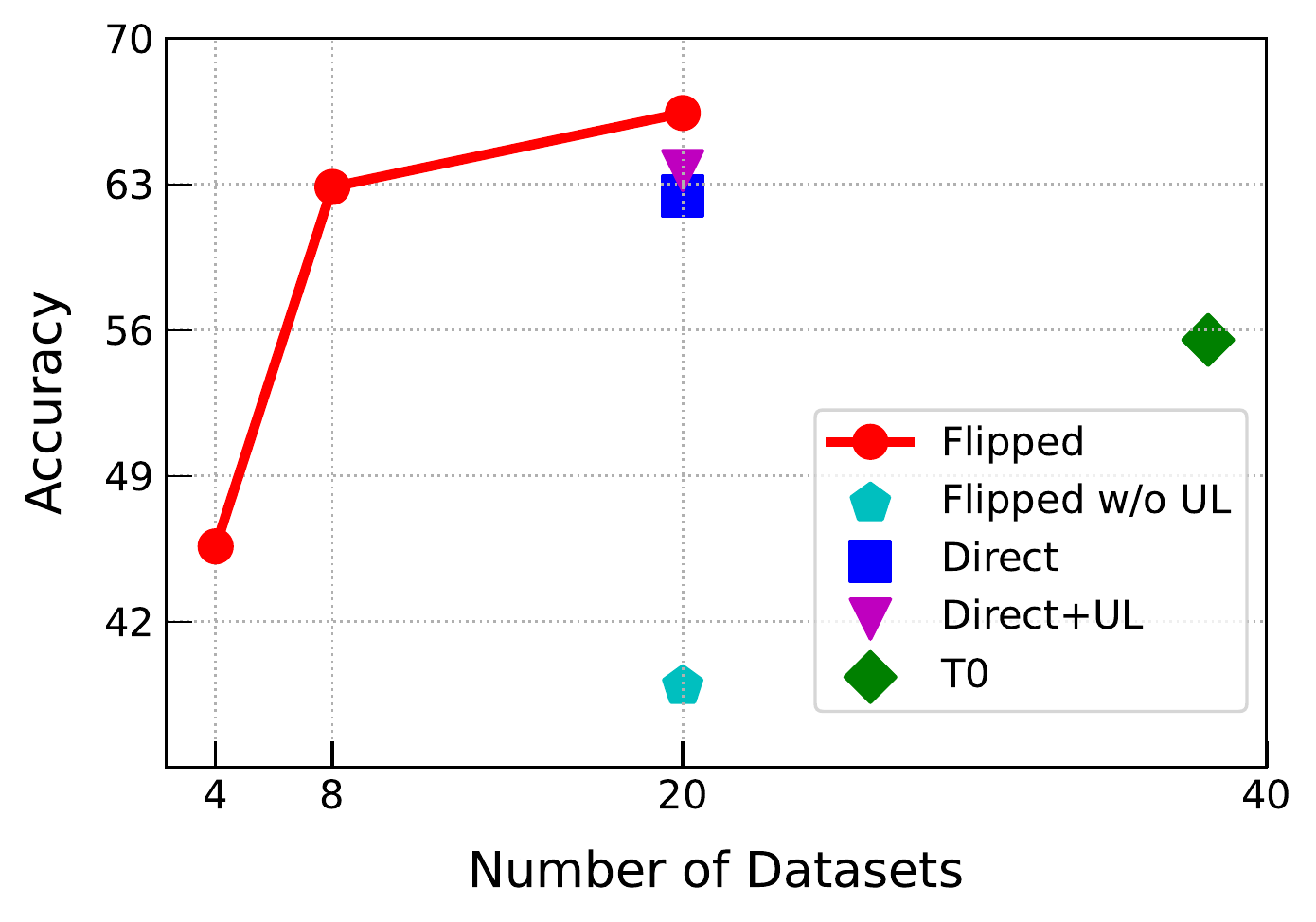}
    \label{fig:query}
    \end{subfigure}
\caption{{\ours trained on varying numbers of datasets. \textsc{Flipped w/o UL} indicates ablation of \ours without unlikelihood training. We also analyze the effect of unlikelihood training on \textsc{Direct} (\textsc{Direct+UL}). \textbf{Left}: Average F1 score of 7 classification tasks. \textbf{Right}: Average accuracy of 7 multi-choice tasks. All models are 3B-sized meta-trained LMs.}}
\label{fig:fig_ablation}
\vspace{-5mm}
\end{figure} 

\subsection{Effect of Unlikelihood Training}

 As mentioned in Section \ref{sec:unlikelihood} and shown in Figure \ref{fig:fig_ablation}, we observe that \training ignores the input instance-label correspondence if unlikelihood loss is not added, hurting the performance significantly. We additionally analyze if the strong task generalization of \ours is solely coming from unlikelihood training by applying unlikelihood training on \textsc{Direct}, which is our strong baseline. As shown in the performance of \textsc{Direct+UL} in Figure \ref{fig:fig_ablation}, unlikelihood training worsens the task generalization performance especially for classification tasks while giving marginal improvement on multi-choice tasks, underperforming \ours for both types of tasks. 
 This shows that the effectiveness of \training is not coming from unlikelihood training itself; both factors of \training, flipping the label and instruction space and unlikelihood training, are needed to generalize effectively on unseen target tasks.

\subsection{Number of Datasets}
\label{subsection:num_dataset}
Meta-trained LMs via direct prompting shows improved performance when the number of datasets increases \citep{sanh2021multitask, wang2022benchmarking, wei2021finetuned}. We also analyze if this phenomenon holds for \training by varying the number of datasets per task cluster; we increase the total number of datasets by 4, 8, and 20. As shown in Figure \ref{fig:fig_ablation}, the performance of \ours increases as the number of datasets increases, similar to LMs trained through direct prompting. Interestingly, using only 8 datasets to meta-train \ours also shows strong performance, outperforming \textsc{Direct} model trained with 20 datasets on multi-choice tasks. Also, this efficient but effective model significantly outperforms T0-3B, while only using 20\% of the number of datasets and 5\% token updates. This shows that \training can result in generalization to unseen tasks while using only a few number of datasets, making not only effective but also \textit{efficient} zero-shot learners.

\section{Additional Experiments}
\begin{figure}[t]
    \centering
    \begin{subfigure}[b]{0.4\textwidth}
    \includegraphics[width=\textwidth]{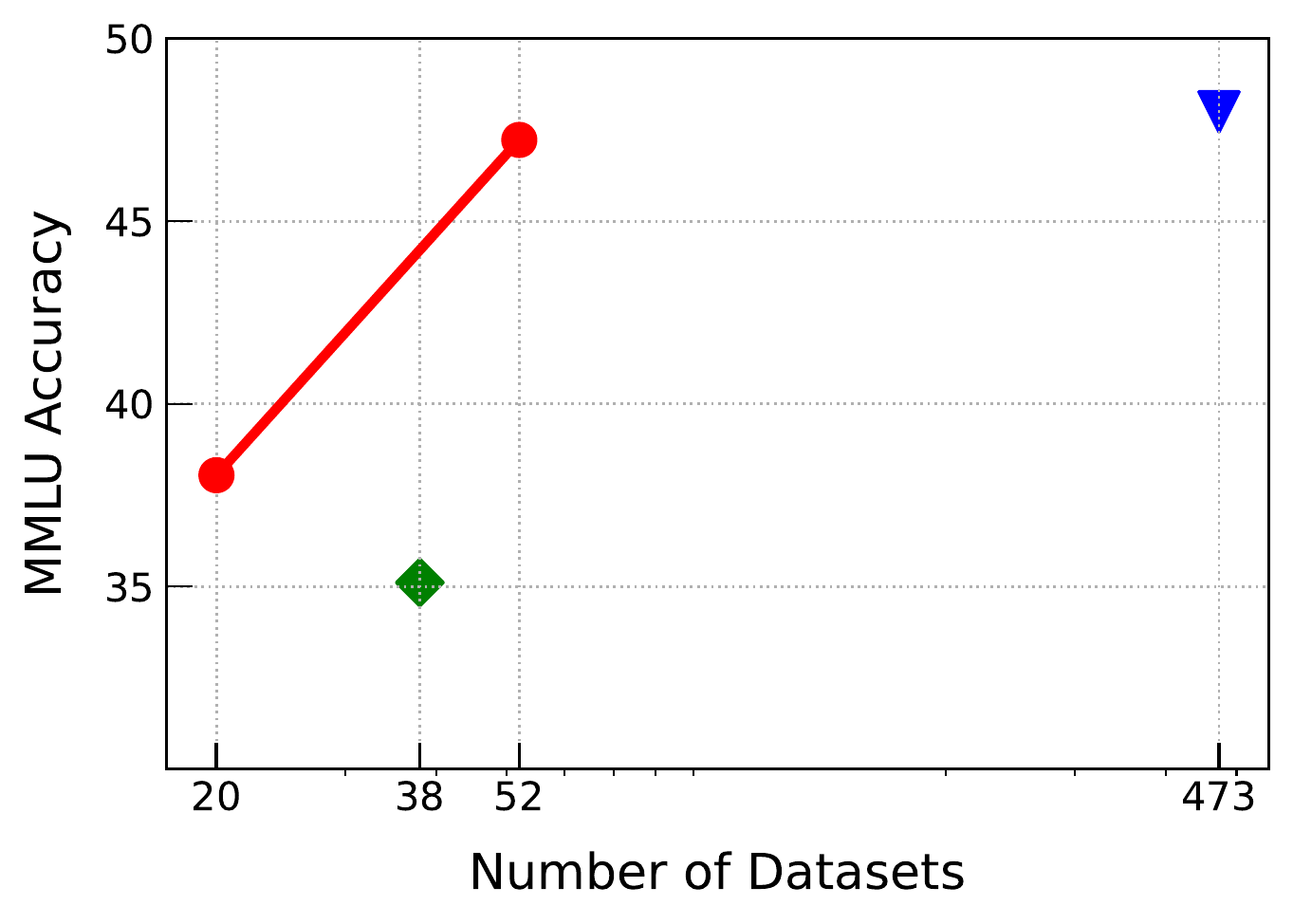}
    \label{fig:data_inst}
    \end{subfigure}
    \hspace{5mm}
    \begin{subfigure}[b]{0.4\textwidth}
    \includegraphics[width=\textwidth]{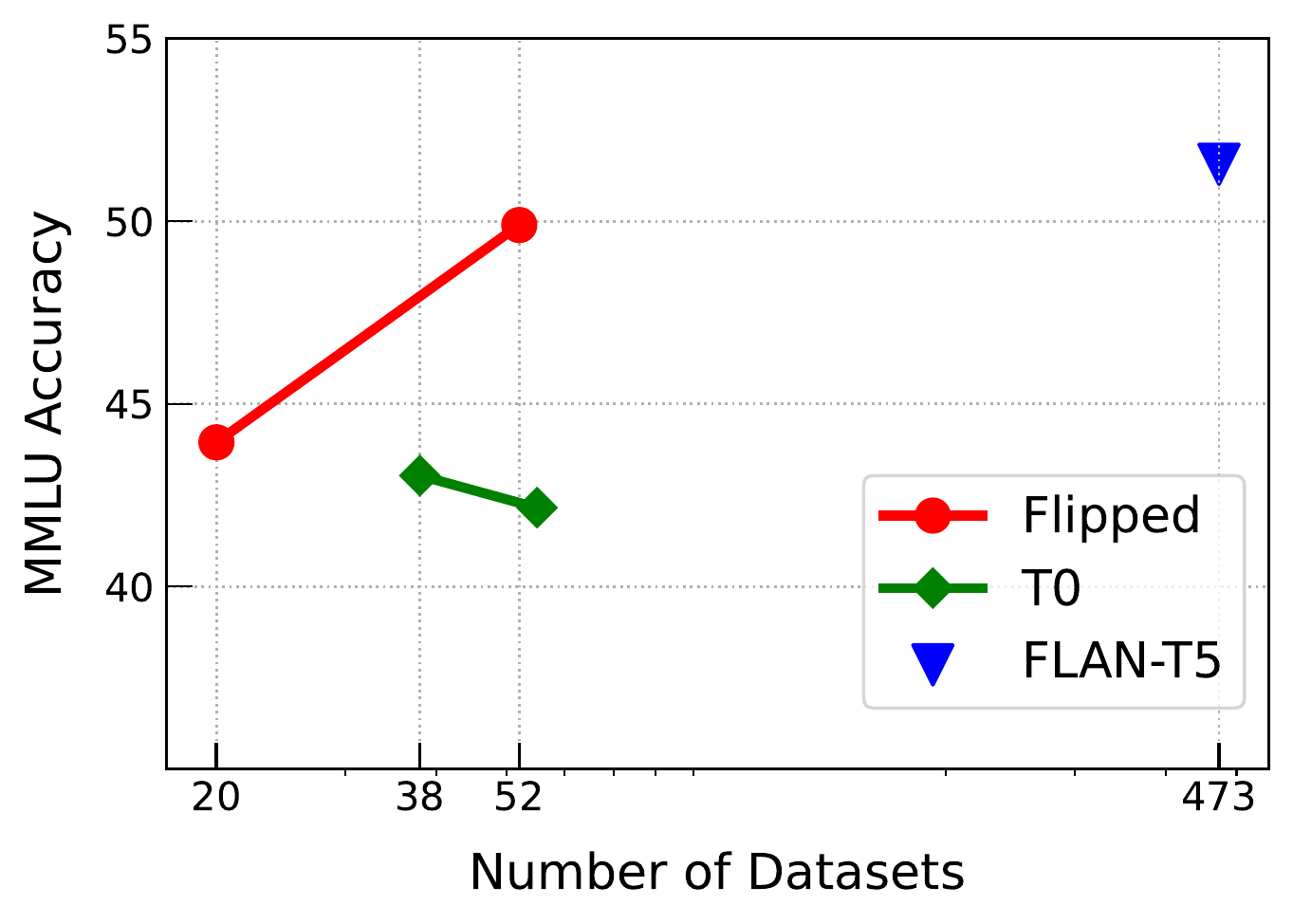}
    \label{fig:query}
    \end{subfigure}
\caption{{Zero-shot MMLU accuracy when scaling the number of datasets during meta-training. FLAN-T5 \citep{chung2022scaling} is trained on 473 datasets (1,836 tasks). Although \ours trained with 52 datasets (\textsc{Flipped+}) uses about 10\% of training datasets compared to FLAN-T5, it largely reduces the performance gap. \textbf{Left}: Average accuracy of 3B-sized models on MMLU benchmark. \textbf{Right}: Average accuracy of 11B-sized models on MMLU benchmark.}}
\label{fig:fig_scaling_task}
\vspace{-5mm}
\end{figure}

Concurrent work of \cite{chung2022scaling} show that scaling the number of training datasets (up to 473 datasets) during meta-training results in state-of-the-art performance on challenging tasks such as the MMLU benchmark \citep{hendrycks2020measuring}. From the findings of Section \ref{subsection:num_dataset}, we also expect that scaling up the number of datasets during meta-training can improve the performance further. Similar to the approach of \cite{chung2022scaling}, we scale up the number of datasets during meta-training by adding generation tasks that are used to train the T0++ model \citep{sanh2021multitask}. For generation tasks, we train with the same training objective as classification tasks. For unlikelihood training of generation tasks, we sample an incorrect label option from a different training instance of the same dataset which is different from the correct label option. The number of training datasets in total is 52 and we refer to the model trained with \training with these datasets as \textsc{Flipped+}. We evaluate \textsc{Flipped+} on the zero-shot setting of the MMLU benchmark and compare the performance with T0 models \citep{sanh2021multitask} and FLAN-T5 \citep{chung2022scaling} on the same model size, shown in Figure \ref{fig:fig_scaling_task}.

Consistent with the result of Section \ref{subsection:num_dataset}, \training additionally benefits from the scale of the number of datasets: \textsc{Flipped+} outperforms \ours for both 3B and 11B sized models. Compared to T0 models, which do not always benefit from scaling the number of datasets, \textsc{Flipped+} shows significant improvement. Moreover, while only using about 10\% of the number of training datasets compared to FLAN-T5, \textsc{Flipped+} largely reduces the performance gap between FLAN-T5. We suggest that using less number of training datasets during meta-training but resulting in strong zero-shot performance is important because it is closer to a \textit{true} zero-shot setting. 

\section{Limitations}
\label{appen:limitations}
In this work, we do not explore \ours for performing unseen tasks that do not have label options such as free-form generation. However, we believe \ours can be used for these tasks as well by obtaining the list of label options from a different LM, which we leave for future work. 
\training also assumes that the task instruction and input instance can be separated during zero-shot inference. However, although instruction-based benchmarks such as Natural Instructions \citep{mishra2021cross, wang2022benchmarking} define the prompted input as a naïve concatenation of task instruction and input instance, this is not guaranteed for prompt libraries such as Promptsource \citep{bach2022promptsource}. Therefore, \training needs additional techniques to separate the task instruction and the input instances as shown in Section \ref{sec:instruct_channel}.

\section{Conclusion}
In this paper, we propose \training, which is a meta-training method that flips the instruction and label space, training the LM to compute the conditional probability of the task instruction given input instance and label. Our findings show that by conditioning on the label space instead of generating it, \training avoids label overfitting, leading to better zero-shot unseen task generalization capabilities especially for tasks that contain various novel labels. To this end, we suggest the research community consider \training for making efficient LMs that can generalize to challenging unseen tasks. 

\subsubsection*{Acknowledgments}
We thank Sewon Min, Sungdong Kim, Miyoung Ko, Seungone Kim, Yujin Kim, and Yejin Cho for helpful discussions. This work was partly supported by Institute of Information \& communications Technology Planning \& Evaluation (IITP) grant funded by the Korea government (MSIT) (No.2022-0-00113, Developing a Sustainable Collaborative Multi-modal Lifelong Learning Framework, 80\%; No.2021-0-02068, Artificial Intelligence Innovation Hub, 10\%; No.2019-0-00075, Artificial Intelligence Graduate School Program (KAIST), 10\%).

\bibliography{iclr2023_conference}
\bibliographystyle{iclr2023_conference}

\appendix
\clearpage



\section{Illustration of Denoising Objective}
\label{appen:denoise_ill}
\begin{figure}[t]
    \centering
    \includegraphics[width=\linewidth]{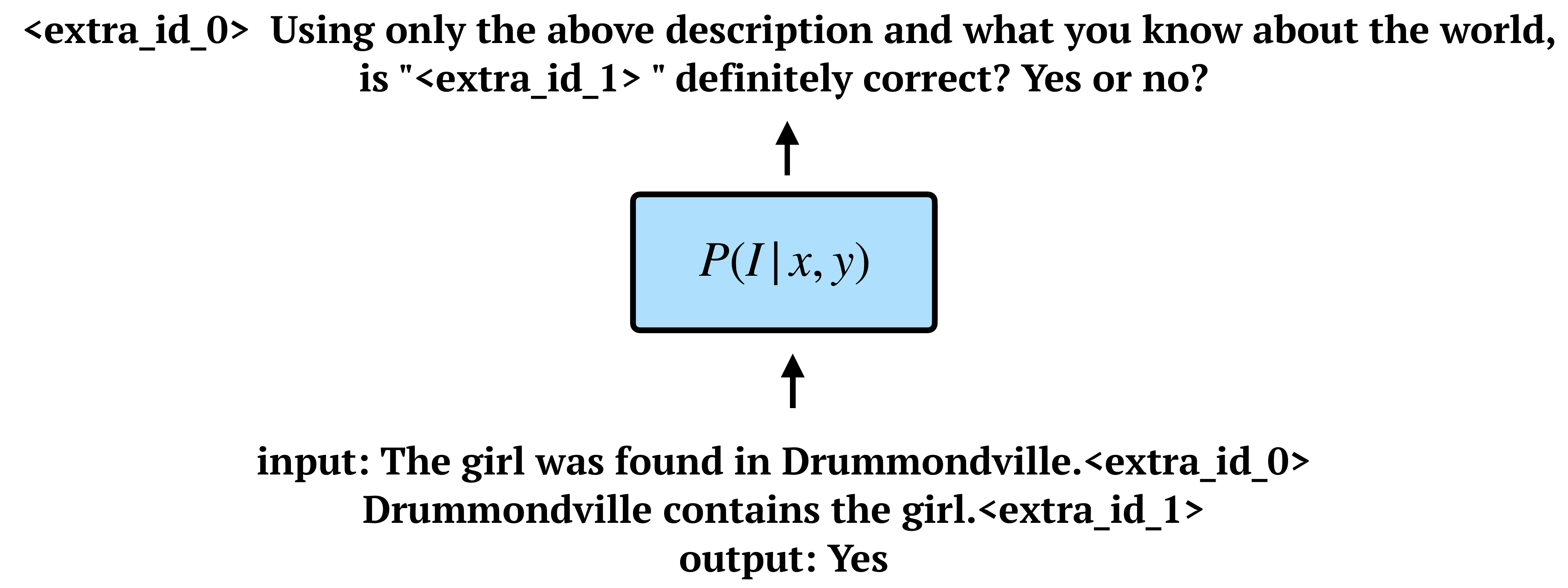}
    \caption{Illustration of denoising objective of \training. Given, an input instance with sentinel tokens, \training makes the LM generate the task instruction corresponding to the sentinel tokens for a correct label option.}
    \label{fig:denoising} 
\end{figure}

As shown in Figure \ref{fig:denoising}, \training uses a denoising objective while meta-training to effectively separate the prompted input obtained through Promptsource \citep{bach2022promptsource} into task instruction and the input instance. By replacing task instruction as sentinel tokens, \training makes the LM generate the task description that corresponds to the sentinel tokens.

\section{Training Configurations}
\label{appen:config}
For backbone LM of \ours, we use T5.1.1 \citep{raffel2019exploring} which is pre-trained on a \textit{denoising} objective while we use T5-LM adapted model \citep{lester2021power} for \textsc{Direct} and \textsc{Channel} which is continually trained T5.1.1 model on \textit{language modeling} objective for 100B additional tokens. We use a different backbone LM for \training because the meta-training objective is denoising objective while \textsc{Direct} and \textsc{Channel} is language modeling objective. From preliminary experiments, we observe that the language modeling training objective of \textsc{Direct} and \textsc{Channel} on T5.1.1 model leads to poor performance. Also, denoising objective of \training on T5-LM adapted model leads to poor performance. Following \citet{sanh2021multitask, raffel2019exploring}, we limit the number of data instances for each dataset to 500,000 to resolve data instance imbalance during meta-training. We train each model for 5K steps, with a batch size of 240. We set input and output sequence lengths as 512 and 128 respectively for \ours-3B. For \ours-11B, we set input and output sequence lengths as 384 and 64 respectively for computational efficiency. For \textsc{Direct} and \textsc{Channel}, we set the learning rate as 1e-4 and for \ours, we set the learning rate as 5e-5 because the training objective is different (generation vs denoising). We set the weight hyperparameter of likelihood and unlikelihood loss as $\lambda=3$. Note that our total training compute used during meta-training is around 5\% that of the training compute used to train the original T0: different from \citet{sanh2021multitask} which uses the batch size of 1024, sequence length of 1024, training steps of 12,200, we use a batch size of 240, half of the sequence length, training steps of 5,000 leading to 4.8\% token updates compared to T0.  For \textsc{Flipped+}, we almost keep the training configurations of \ours with only a few variations. Unlike \ours, we limit the number of data instances for each dataset to 50,000 to resolve data instance imbalance during meta-training. Also, for 3B-sized \textsc{Flipped+}, we train for 10K steps during meta-training due to the increased number of datasets. For 11B-size \textsc{Flipped+}, we keep the number of training steps to 5K steps due to computational costs.

\section{Training and Evaluation Datasets}

\subsection{Meta-training Datasets}
\label{appen:train_datasets}
We use 4 task clusters for meta-training of \textsc{Direct}, \textsc{Channel} and \ours: sentiment classification, paraphrase, topic classification, which is 20 datasets in total. We use imdb \citep{maas-etal-2011-learning}, amazon\_polarity \citep{McAuley2013HiddenFA}, rotten\_tomatoes \citep{pang-lee-2005-seeing}, yelp\_review\_full \citep{zhang2015character}, app\_reviews for sentiment, glue/qqp \citep{2018arXiv180407461W}, paws/labeled\_final \citep{zhang-etal-2019-paws}, glue/mrpc \citep{dolan-brockett-2005-automatically} for paraphrase, ag\_news \citep{NIPS2015_250cf8b5}, dbpedia\_14 \citep{Lehmann2015DBpediaA} for topic classification, cos\_e/v1.11 \citep{rajani2019explain}, dream \citep{sun-etal-2019-dream}, quail \citep{Rogers_Kovaleva_Downey_Rumshisky_2020}, quartz \citep{tafjord-etal-2019-quartz}, social\_i\_qa \citep{sap-etal-2019-social}, wiqa \citep{tandon-etal-2019-wiqa}, cosmos\_qa \citep{huang-etal-2019-cosmos}, qasc \citep{khot2020qasc}, quarel \citep{tafjord2019quarel}, sciq \citep{welbl-etal-2017-crowdsourcing} for multi-choice QA.

\subsection{Evaluation Datasets}
\label{append:eval_datasets}
We evaluate on 14 datasets of BIG-bench benchmark \citep{srivastava2022beyond}: Known Unknown, Logic Grid, StrategyQA, Hindu Knowledge, Movie Dialog, Code Description, Conceptual, Language ID, Vitamin C, Syllogisms, Misconceptions, Logical Deduction, Winowhy, Novel Concepts, following \citet{sanh2021multitask}. For English NLP tasks, in addition to 11 unseen evaluation datasets from \citet{sanh2021multitask}, we add 3 unseen question-answering datasets from \citet{lin2022unsupervised}, resulting in 7 classification (RTE \citep{dagan2005pascal}, CB\citep{de2019commitmentbank}, ANLI R1,R2,R3 \citep{nie2019adversarial} WSC \citep{levesque2012winograd}, WiC \citep{pilehvar2018wic}) and 7 multi-choice datasets (COPA \citep{roemmele2011choice}, Hellaswag \citep{zellers2019hellaswag}, Storycloze \citep{mostafazadeh2016corpus}, PIQA \citep{bisk2020piqa}, ARC-Challenge \citep{clark2018think}, OpenbookQA \citep{mihaylov2018can}). We exclude SQuAD2.0 which is included in evaluation setting of \citet{lin2022unsupervised} because it does not have label options.

\section{Evaluation Setting}
\label{appen:evaluation}
For the result of PaLM and GPT-3 of Table \ref{table:main}, we use the performance reported in the paper. For the result of GPT-3 on zero-shot setting in Table \ref{table:nlp}, we use the performance reported in the paper for multi-choice tasks while we rerun the experiments using OpenAI API for classification tasks to report F1 scores. We used the prompt named `GPT-3 style' for every dataset of Promtpsource library. For experiments of Figure \ref{fig:label_generalize}, we randomly sample 1,000 data instances for seen task label generalization evaluation,  for efficiency.


\begin{table}[]
    \centering
    \begin{tabular}{l|cc}
    \toprule
     & Classification & Multi-choice \\ \midrule
    T0-3B & 36.79 & 55.53 \\
    T0-3B + Calibration & 33.59 &46.40 \\
    \ours & \textbf{44.95} & \textbf{66.43}
    \end{tabular}
\caption{Effect of calibration on T0-3B meta-trained LM. Results show that the performance worsens if calibration is applied especially for multi-choice tasks.}
\label{table:calibrate}
\end{table}

\section{Calibration Results}
\label{appen:calibrate}
Previous work has used calibration methods to match the label distribution of the target task during inference of zero-shot setting \citep{zhao2021calibrate, holtzman2021surface}. We also analyze if calibration is effective for meta-trained LMs by applying contextual calibration on T0-3B. Because we evaluate the zero-shot task generalization performance, we use the probability of the label given an empty string for calibration. As shown in Table \ref{table:calibrate}, applying calibration \textit{hurts} the performance of meta-trained LMs.

\section{Label Pair Variations}
We provide the list of variations of label pairs on Table \ref{table:binary_variation} and Table \ref{table:multi_variation}. Table \ref{table:binary_variation} shows the label pair variation of binary classification datasets (RTE, WiC, IMDB, PAWS) while Table \ref{table:multi_variation} shows the label pair variation of CB, which consists of 3 label options.
\label{appen:label_20}
\begin{table}[]
\centering
\renewcommand{\arraystretch}{1.2}
\begin{tabular}{c|c}
\midrule
yes & no \\
true & false \\
positive &negative \\
right & wrong \\
correct & incorrect \\
agree & disagree \\
good & bad \\
guaranteed & impossible \\
always & never \\
affirmative & contradicting \\
exactly & not ever \\ 
undoubtedly & not at all \\
fine & disagreeable \\
good enough & cannot be \\
definitely & never \\
unquestionable & no way \\
yep & nope \\
yea & nah \\
without doubt & refused \\
willing & unwilling \\ \midrule
\end{tabular}
\caption{List of 20 pairs of labels used to evaluate label generalization on binary classification datasets (RTE, WiC, IMDB, PAWS).}
\label{table:binary_variation}
\end{table}

\begin{table}[]
\centering
\renewcommand{\arraystretch}{1.2}
\begin{tabular}{c|c|c}
\midrule
yes & no & maybe \\
true & false & neither \\
positive & negative & inconclusive \\
right & wrong & perhaps \\
correct & incorrect & might be \\
agree & disagree & could be \\
good & bad & neutral \\
guaranteed & impossible & possible \\
always & never & sometimes \\
affirmative & contradicting & feasible \\
exactly & not ever & as it may be \\
undoubtedly & not at all & doubtfully \\
fine & disagreeable & conceivable \\
good enough & cannot be & can be \\
definitely & never & uncertain \\
unquestionable & no way & questionable \\
yep & nope & iffy \\
yea & nah & nn \\
without doubt & refused & controversial \\
willing & unwilling & not for sure \\ \midrule
\end{tabular}
\caption{List of 20 pairs of labels used to evaluate label generalization for CB, which has 3 label options.}
\label{table:multi_variation}
\end{table}

\end{document}